# Searching for Image Information Content, Its Discovery, Extraction, and Representation

Emanuel Diamant
VIDIA-mant, POB 933, Kiriat Ono 55100, Israel
emanl@012.net.il

**Abstract**

Image Information Content is known to be a complicated and a controversial problem. This paper posits a new image information content definition. Following the theory of Solomonoff - Kolmogorov - Chaitin's Complexity, we define image information content as a set of descriptions of image data structures. Three levels of such description can be generally distinguished: 1) the global level, where the coarse structure of the entire scene is initially outlined; 2) the intermediate level, where structures of separate, non-overlapping image regions usually associated with individual scene objects are delineated; and 3) the low level description, where local image structures observed in a limited and restricted field of view are resolved. A technique for creating such image information content descriptors is developed. Its algorithm is presented and elucidated with some examples, which demonstrate the effectiveness of the proposed approach.

## 1. Introduction

In recent years, we have witnessed an explosive growth of visual information in our surrounding environment. To access it, to retrieve and efficiently manipulate it, one needs appropriate image-processing tools and utilities. It is only natural to expect that they will be capable of handling images in a human-like manner, that is, in accordance with image-centric information. Progress in meeting such requirement is hindered by lack of specific knowledge about human vision system properties (as well as lack of fully utilizing what is already known of such properties) and the ways in which effective image processing is actually accomplished by humans. Such lack of knowledge results in various ad hoc solutions, approximations, and misconceptions. One such misunderstanding is that image information is inseparable from image data. In lack of appropriate means for image information extraction and management we are forced to handle it in the most primitive and ineffective way – by means of raw image data. That is, by merely the code values attached to the pixels.

Another common misunderstanding is that image semantics is essentially what we are looking for in an image. It is well known that different people can perceive one and the same image differently. On the other hand, "perceptual illusions" and various kinds of "perceptual blindness" are common for human vision practice. This means: we see what does not really exist in an image and we do not see what it actually contains. Image information is substituted with our impression about it or with our interpretation of it. This makes visual information a sort of "virtual reality".

The deliberate objective of this paper is to promote a view that Image Information Content is an intrinsic property of an image, stand alone and objective. That is, it does not depend on any subjective (human) interaction with it and is determined solely by the available image data. We suggest the following definition: Image Information Content is a set of recursive descriptions of discernable image data structures perceived at different visibility levels. Assuming that the descriptions are created with a syntactically defined and fixed language, the total length of the descriptors may be considered as a quantitative measure of the image contained information.

## 2. Description of Image Contained Information

Our approach to the problem of image information content discovery and representation stems from the basic principles of Kolmogorov Complexity, also known as Inductive Inference Theory, or as Algorimic Information Theory, or as Kolmogorov-Chaitin Randomness. Such diversity in names is due to the fact that similar ideas were introduced independently (and more or less at the same time) by R. Solomonoff (1964) [1], A. Kolmogorov (1965) [2], and G. Chaitin (1966) [3]. In such a case, a name like "Solomonoff-Kolmogorov-Chaitin Complexity" would be more suitable (to give proper credit to all of the inventors). But the name "Kolmogorov Complexity" has become far more widely and frequently used. So, following the general preference, we shall also use it in the subsequent discussion.

Kolmogorov Complexity is a mathematical theory devised to explore the notion of randomness. Its basic concept is that information contained in a message (an image can be considered as a message) can be quantitatively expressed by the length of a program, that (when executed) faithfully reproduces the original message, [4]. Such a program is called the message description. The original theory deals with binary strings, which are called objects. Although we use "objects" while meaning "image objects" (instead of "binary string objects") and use "object information content description" while meaning "image information content description", such substitutions are tolerable, because any object can be described by a finite string of signs ("letters") taken from some fixed finite alphabet. By encoding these letters in bits (1's and 0's) we can reduce every object description to a finite binary string, [5].

Various description languages can be devised and put to use for the purpose of description creation. It is only natural to anticipate that a specific language will influence the length of the description and its accuracy. One of the important finding provided by Kolmogorov complexity theory is the notion of language invariance, [5]. That is, the description language, of course, affects the length of object's description, but this influence can be taken into account by a language dependent constant added to the body of a language independent description. This language independent description actually is the Kolmogorov complexity of an object. It determines the unconditional amount of information in an individual object, and thus can be called the absolute Kolmogorov complexity, [4]. The problem, however, is that this absolute Kolmogorov complexity is (theoretically) unconstrained and, thus, it is (practically) uncomputable.

We shall in a moment return to the implications of this important issue, but now we would like to proceed with further examination of the Kolmogorov complexity insights. To ensure the effectiveness of object description, the theory specifies that the shortest object description must be contrived at first. An important equivalence between the shortest object description and the simplest object structure is then established, [6]. This, in turn, induced further developments (introduced to the theory by Kolmogorov and by his students, coworkers and followers), such as the notion of a two-part code [7], the Kolmogorov's Structure function [8], the "Sophistication" and "Meaningful Information" notions [9]. They may be summarized as follows: To get the most effective description code its creation must commence with the description of the most simple object's structure. The best way to achieve an object simplification is some sort of object compression, when the existing object regularities are simply squeezed out from it [7]. The remaining part of the object data, the structure of which was not captured by the first step, should be processed in the same manner – that is, the regularities discernable at the next simplification level must be squeezed out and the structures observable at this level must be described (encoded), [8]. A hierarchical and recursive strategy for a description creation is thus evolved: Beginning with the simplified and course object structure, the description is subsequently augmented with more and more fine details unveiled at different (lower) hierarchical levels of object analysis and description.

Here we return to the problem of unattainable low-level information content descriptors. Practically that means that some part of object data would always remain undiscovered and undescribed, [5]. This would sound less discouraging if we reinstate our primary practical goal – to discover, extract and represent the relevant image contained information. Indeed, not all of the available visual information is actually needed and used in common image-processing tasks. The required level of information details is always determined by user's intention or by the specific tasks at hand. To address this peculiarity and to gain the necessary information details, we usually resort to changes in their (details) visibility, as we do with microscope and telescope instruments.

Therefore, we can put aside the problem of description level ambiguity, and accept the insights of Kolmogorov Complexity theory as a paradigm for image information content discovery, extraction and representation. Consistent with this goal, our proposed approach may be specified as:

- Image information content is a set of descriptions of observable image data structures.

- These descriptions are executable, that is, following them the meaningful part of image content can be faithfully reconstructed.

- These descriptions are hierarchical and recursive, that is, starting with a generalized and simplified description of image structure they proceed in a top-down fashion to more and more fine information details resolved at the lower description levels.

- Although the lower bound of description details is unattainable, that does not pose a problem for an effective description creation, because information content comprehension is underserved by fine details.

- In exceptional cases, when detail deprivation is inappropriate, the two-part code principle of Kolmogorov complexity can be adopted as a solution.

## 3. Biological Vision Supporting Evidence

The above approach to image information content comprehension significantly deviates from the traditional and currently used approaches, which assume information content in the Shannon's sense (that means, averaged over the whole picture space, and not object-related as in the Kolmogorov's sense). In this case, a search for some supporting concepts from biological vision would be appropriate.

Currently dominating theories of visual perception basically assume that human visual system starts image information processing at very low processing levels, where primitive local features (such as edges, color, texture or depth information) are resolved and located within an image. These features are then arranged in larger and more complex agglomerations, which are fed into higher brain centers for further generalization, resulting in a faithful and accurate image perception. In most contemporary theories, the flow of information processing proceeds strictly in the bottom-up direction and therefore the approach is known as such – the "bottom-up approach". Its roots can be traced back to Treisman's Feature Integrating Theory [10] or Biederman's Recognition-by-components theory [11]. Its principal presupposition is that visual properties of an image are simultaneously available and can be processed in parallel across the entire visual field. Although never explicitly certified, that implies that the whole visual field is spatially uniformly sensed and is presented to the processing system like a pixel array of a CCD camera. If you know from your own experience that your mental image of the external world is always unexceptionally acute and sharp, such an assumption will sound to you quite normal and natural.

However, real things look somewhat different. The eye's retina has an arrangement where only a small fraction of the view field (approximately $2^0$ out of the entire field of $160^0$, [13]) is densely populated with photoreceptors. Only this small zone, called the fovea, is responsible for our ability to see the world sharp and clear. The rest of the view field is a fast descending (in spatial density) placement of photoreceptors (from the fovea outward to the eye's periphery), which provides the brain with crude and fuzzy impression about the outside world. The initial attempts to explain this arrangement assumed that the continual eye movements (also known as saccades) might be to compensate for the lack of the resolution over the visual field. Eye saccades, thus, for many years became an object of scrutiny, and not the peripheral vision functionality.

Another assumption was that eye saccades are primarily compensating for the decaying nature of neurons' responses, due to their adaptation to constant light illumination. The saccade movement has been seen as a countermeasure against image fading in a fixed eye position and as means of continuous image refreshment, ([12], p. 253). Neither of these explanations challenged the basic presupposition that a complete sensory image is retained across each saccade and is fused with the new sensory image at every eye fixation. The extensive research work in the field of visual attention, which elucidate saccade movement as the prime mechanism for efficient information flow management (when limited processing resources are sequentially directed to the most prominent image parts), did not change these assumptions. According to attentional vision theories, high-resolution (low-level) image details are resolved after the decision to make a saccade is accomplished and only when the fovea is fixed over a particular (deliberately selected) image location. This contradicts the traditional premise that low-level feature detection and verification precedes image analysis and decision making. Nevertheless, the most elaborated theory of attention selection, which is based on an idea of a master saliency map, builds up its intermediate feature maps in a bottom-up fashion [14, 15].

At the same time, accumulating evidence from psychophysical studies have shown that primate visual system can analyze complex visual scenes in extremely short time intervals, practically "at a glance", and that the first signals reaching the brain are from the eye's periphery, [16]. This inevitably led to erosion and revision of the dominant bottom-up approaches, (which do not permit such short processing times). Schyns and Oliva have made a suggestion that visual recognition/categorization tasks can use "express", but comparatively imprecise and coarse-scale representations, before the fine-scale representations are acquired, as early as in 1994, [17, 18]. (Although they themselves reference to an even more earlier work of Henderson dated 1992, [19].) Thereafter many publications [20-22] have continued to investigate and to elaborate this issue, with results that "are in keeping with other findings from humans that global properties are sometimes perceived better than local ones and thus might be basic", [23]. Such findings in biological vision favor top-down input as opposed to the traditional bottom-up approach. This seems to be a more suitable way of image information content perception and comprehension.

The term "top-down" processing is ubiquitously used in traditional approaches. The accepted meaning is that to facilitate appropriate feature selection, some external knowledge (usually related to the task at hand or to user-defined requirements) must be incorporated into the process. Such knowledge usually comes from the higher (cognitive) brain levels and that is the reason for the top-down intervention (or mediation) into the primary bottom-up process.

The use of "top-down" in this paper implies that information unfolds from its most coarse, crude and simplified form (at the top representation level, which is the lowest resolution level) to more specified, enriched and detailed representation at lower levels. This process is totally independent from any positioned cognitive higher-levels as well as from the task at the hand or user's expectations (that was already declared previously in the Section 1). Obviously, their impact cannot and must not be denied. But the interaction with them must take place in another form, presumably outside the top-down processing path of our approach.

One final remark: Biological vision was always an unlimited source of inspiration to computer vision designers. However, this fruitful interplay is packed with misunderstandings. One of the most popular is that the eye sees the surrounding world like a CCD camera – with a high and uniform resolution over the entire field of view. For almost 40 years of computer vision, image analysis was pixel-centered and image-understanding processing was bottom-up directed – from pixels to features, from features to objects, from objects to a general scene perception. In the light of recent biological vision findings such as that our conscious experience of a complete visual scene is an illusion, [22], and that the top-down user-independent information unfolding in the visual cortex is a fact, a shift away from pixel-centric approaches to new image processing approaches is certainly expected in computer vision.

## 4. Creating Higher Level Information Content Descriptions

### 4.1 General Considerations

In the previous sections we have outlined a general scheme for image information content discovery, extraction and presentation. Now we would like to describe some implementation details for an algorithm that accomplishes the proposed concepts. Its architecture is shown in Figure 1, and it is comprised of three main processing paths: the bottom-up processing path, the top-down processing path, and a lane where the generated descriptors are actually accumulated (in a diversity of their representation forms – alphanumeric textual descriptors, parameter lists, localization maps, etc.).

### 4.2 Implementation issues

According to Kolmogorov Complexity definitions, description creation must commence with depiction of the most general and abstracted (simplified) data structures that can be distinguished within an image. To facilitate this requirement, we use a hierarchy of multi-level multi-resolution image representations called multi-stage image pyramid [24]. Such pyramid construction generates a set of compressed copies of the original input image. Each image in the sequence is represented by an array that is half as large as its predecessor. This is usually called a Reduce operation. Its rules are very simple and fast: four non-overlapping neighbor pixels in an image at level $L$ are averaged and the result is assigned to a pixel in a higher ($L+1$)-level image. This is known as "four children to one parent relationship". Mathematically it looks like this:

$$g^{l+1}(x,y) = \left[g^{l}(2x,2y) + g^{l}(2x+1,2y) + g^{l}(2x+1,2y+1) + g^{l}(2x,2y+1)\right]/4 \qquad (1)$$

Where $g^{l+1}(x,y)$ is the gray level value of a pixel at $(x,y)$ coordinate position in a higher $(L+1)$ level image (the parent), and $g^{l}(2x,2y)$ as well as its three nearest neighbors are the corresponding pixels (the children) within an image array at the lower level $L$.

At the top of the pyramid, the resulting coarse image undergoes a round of further simplification. Several image zones, representing perceptually discernible image fractions (visually dominated image parts, super-objects) are determined (segmented) and identified by assigning labels to each segmented piece. Since the image size at the top of the pyramid is significantly reduced and since in the course of the bottom-up image squeezing a severe data averaging is attained, the image segmentation/classification procedure does not demand special computational resources. Thus, any well-known segmentation methodology will suffice. We use our own proprietary technique that is based on low-level information content discovery process (which is described later and also in [25]).

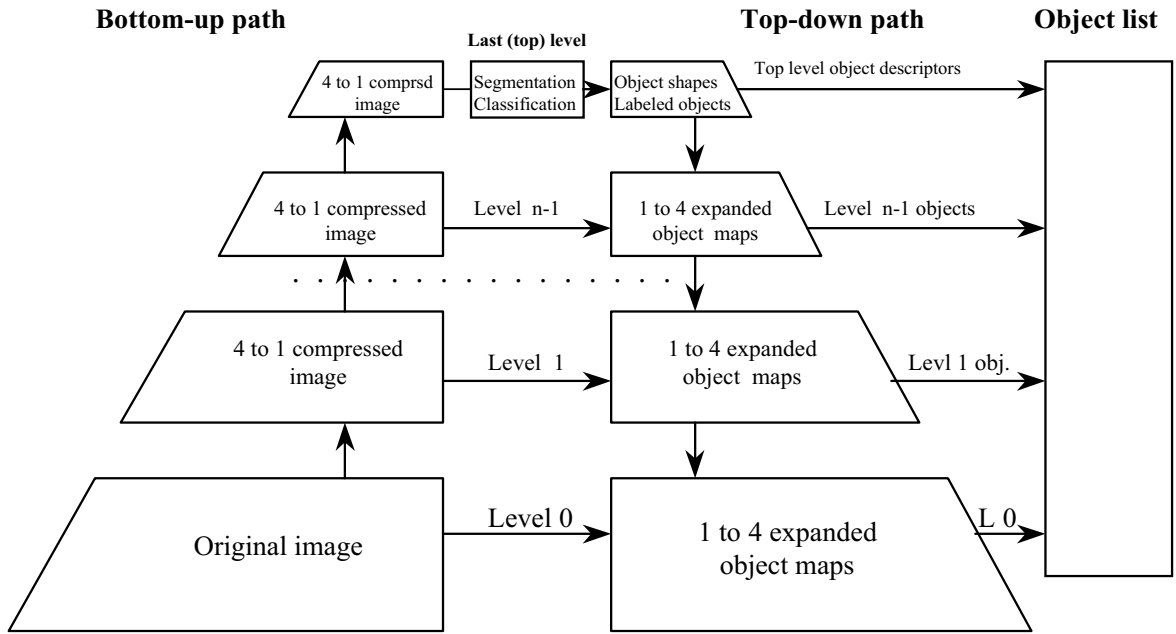

Fig. 1. Block Diagram of the proposed higher-level descriptors creation.

The technique first outlines the borders of the principal image fragments. Then similarly appearing pixels within the borders are aggregated in compact, spatially connected regional groups (clusters). Afterwards, every cluster is marked with a label. Thus, a map of labeled clusters, corresponding to perceptually discernible image regions, is produced. Finally, to accomplish top-level object identification for each labeled region, its characteristic intensity is computed as an average of labeled pixels. This way, a second (additional) segmentation map is produced, where regions are represented by their characteristic intensities.

From this point on, the top-down processing path is commenced. At each level, the two previously defined maps are expanded to the size of the image at the nearest lower level. The expansion rule is very simple: the value of each parent pixel is assigned to its four children in the corresponding lower level map (a reversed Reduce operation). Since the objects at different hierarchical levels do not exhibit significant changes in their characteristic intensity, the majority of newly assigned pixels are determined in a sufficiently correct manner. Only pixels at object borders (and seeds of newly emerging objects) may significantly deviate from the assigned values. Taking the corresponding current-level image as a reference (the left side, bottom-up image in the scheme), these pixels can be easily detected and subjected to a refinement cycle. Here they are allowed to adjust themselves to the "proper" nearest neighbors, which certainly belong to one of the previously labeled regions (or to the newly emerging ones). The process resembles image reconstruction in the Burt-Adelson's Laplacian Pyramid [26], however in our case the exact reconstruction of an image is not required. Our goal is just a simplified image approximation. In such a manner, the process is subsequently repeated at all descending levels until the segmentation/classification of the zero-level (original input image) is successfully accomplished.

It is clear that the reconstructed image is not a "Just Notified Distortion" version of the original one. However, for most decision-making purposes (and that is the primary goal of the biological and machine vision systems) an exact and detail-preserving image information content description is irrelevant. The desired level of information details is task and user intention dependent. Moreover, at different levels of task planning different levels of information abstraction are essential, and they are frequently interchanged during even simple task execution. Only in special cases (medical, scientific, military, fine-art and a couple of other applications) the reconstruction fidelity of the original image can be critically important. In such cases the two-part coding principle of Kolmogorov complexity can be successfully implemented. This means, the information contained in the structures is encoded (described) "as usual", the random (indescribable) part is preserved and attached as a supplementary residual.

The third constituting part of the scheme is the objects' appearance list, where every image object at every processing level (just recovered or an inherited one) is registered. The registered object parameters are the available simplified object's attributes, such as size, center of mass position (coordinates), average object intensity and hierarchical and topological relationship within and between the objects ("sub-part of…", "at the left of…", etc.). They are sparse, general, and yet specific enough to capture the object's characteristic features in a variety of descriptive forms.

We suggest this part of the processing scheme is the most suitable and natural place for external user interaction (a place for the "classical" top-down interference). User-defined task-dependent requirements can be formulated in human-friendly and human-accustomed forms, which are provided and supported by the description languages used. The desired levels of description details are transparent in such a list and are easily attended.

**4.3 Experimental results**

To illustrate the qualities of our approach, we have chosen an image of a typical city scene and launched the algorithm to decompose it to its constituent objects (perceptually discernable visual regions) and to create a list (a hierarchy) of their descriptions. The original image (in Fig. 2) is the size of 640x480 pixels, while the usual image size at the top processing level is approximately 12x12 pixels. This results (in the case of this particular image) in a 6-levels processing hierarchy (where level zero is the original image itself).

The intermediate processing stages are presented in Figures 3 – 7. For observer's convenience, all intermediate (consequential levels) results are rescaled to the size of the original image. From various description forms, accumulated in the Object List, we have chosen only the characteristic intensity maps to be represented in this example. We think they are perceptually close to human's apprehension of image content, and they are well enough suitable to communicate the feeling of consistently growing complexity of image information content description.

The scene complexity grows in the following order: Fig. 3 – Fig. 4 – Fig. 5 – Fig. 6 – Fig. 7. The number of objects identified at each level (and provided in each figure's caption) is given in a cumulative fashion, that is, newly emerged objects are added to the early-identified ones, and the total number of objects (at this level) is presented. This way, the consistency of object labels is preserved via all processing levels.

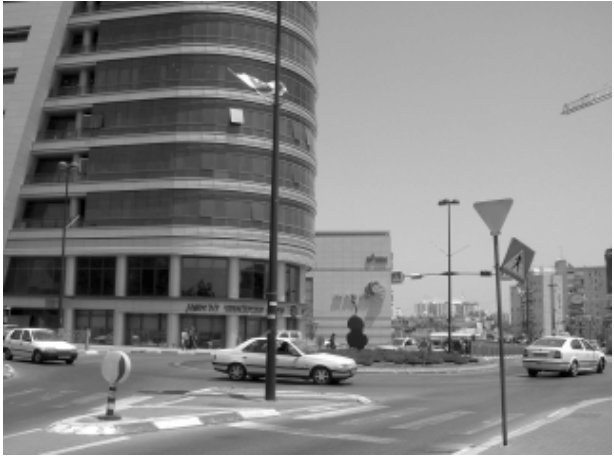

Fig. 2. Original image, size 640 x 480 pixels.

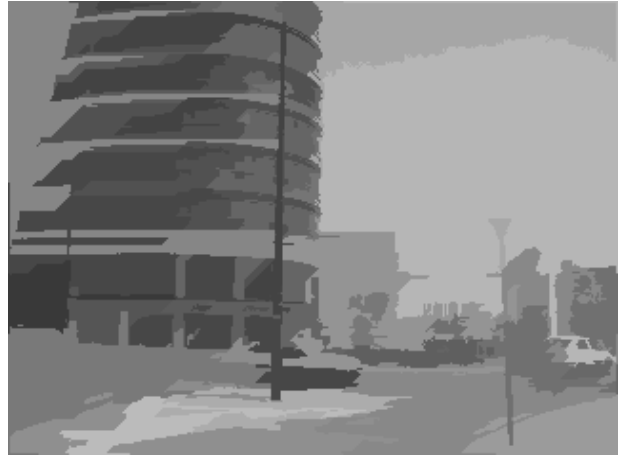

Fig. 3. Level 5 segmentation, 29 objects (regions).

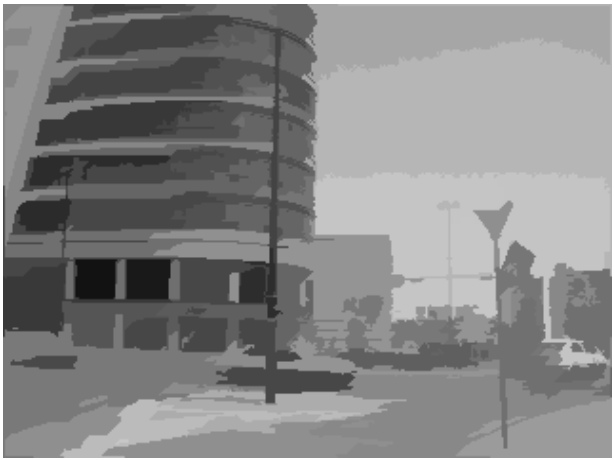

Fig. 4. Level 4 segmentation, 52 objects (regions).

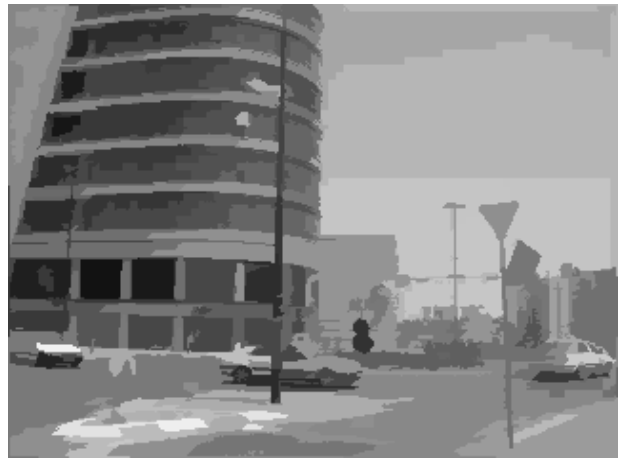

Fig. 5. Level 3 segmentation, 101 objects (regions).

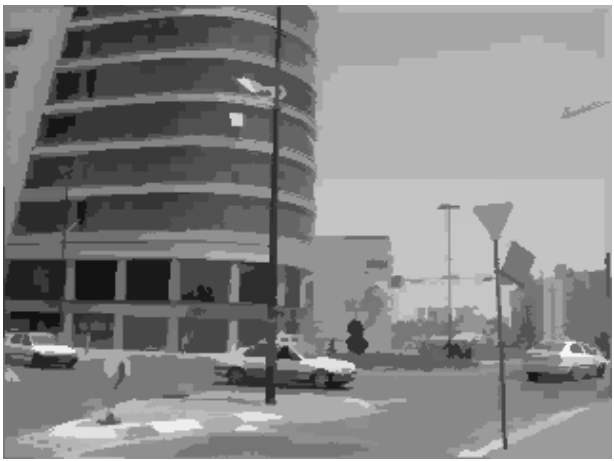

Fig. 6. Level 2 segmentation, 186 objects (regions).

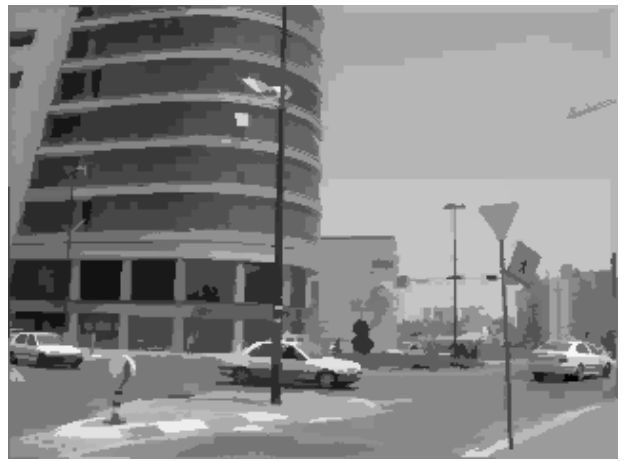

Fig. 7. Level 1 segmentation, 230 objects (regions).

## 5. Creating Low Level Information Content Descriptions

### 5.1 General Considerations

As it was already mentioned, for a long time low-level image features have been the first and the prime core of interest, both in biological and in computer vision. Many relevant low-level-processing mechanisms have been developed during the past years; many of them have been significantly reconsidered and improved with the advent of attentional vision knowledge. According to the established approaches, low-level (local) image information is represented by a feature vector, which contains a number of units associated with local, spatially restricted interactions between neighboring pixels. Because the only structure that can be perceived at such (local) level is an edge structure, the features are, essentially, descriptions of edge properties (intensity gradient, orientation, etc). The total number of features usually taken into account may be quite impressive. For instance, 42 features are used in [14], 27 features in [27], 48 and 60 features in [28]. That makes the representation (of local information content) cumbersome.

As usual, it is appropriate to take a glance on what is going on in biological vision. Although natural evolution does not always select the best solution and mostly compromises at a just fitting one, it is hard to believe that such a hardly manageable description of local information had been overlooked by natural selection and survived. Indeed, contemporary biological vision studies agree that "this information is coded in a relatively abstract (nonperceptual) format", [29], and "informativeness" is hardly a mixture of attribute features [29]. Supporting evidence slowly emerges. Color-independent shape selectivity is reported in [30], orientation invariance of neuronal responses is registered in [31]. We incorporate this evidence, which basically coincides with our intuitive assumptions.

### 5.2 Implementation issues

According to the notions of Kolmogorov Complexity, information descriptors at any description level must strive for their simplest (shortest) possible form. The idea that (low level) information can be measured and can be expressed as a measurable quantity was first introduced by Shannon, [32]. He also cast the logarithmic form of such expression, because "It is nearer to our intuitive feeling as to the proper measure", [32]. We have "the intuition" that in our case another form of expression (description) may be more appropriate.

As it was already mentioned, the only discernable structure at a restricted spatial location is just a discontinuity in spatial homogeneity, which is usually called an "edge". It is a common understanding that two factors must be taken into account when information content of a local image arrangement is considered. First, a measure of topological confidence (or uncertainty) must be evaluated: To what extent does the spatial organization of a pixel and its surrounding neighbors indeed display an edge structure? We define the first component as topological information $I_{top}(x, y)$. Then, the strength of local intensity discontinuity has to be evaluated – to what extent does the intensity change at a given position influence edge visibility at this location?[1] We denote the second component as the intensity information $I_{int}(x, y)$. Finally, a measure of local information content $-I_{loc}(x, y)-$ may be defined as a product of the two constituting components:

---

[1] The term intensity change is applicable to different properties of local discontinuity. For the sake of our discussion, we consider the changes of pixels' gray level values. But the same is true, for example, for pixels' chromatic properties.

$$I_{loc}(x, y) = I_{top}(x, y) \cdot I_{int}(x, y) \quad (2)$$

Here $(x, y)$ are pixel's coordinates in an image array. To decrease the computational burden, here and further we assume the simplest local spatial organization for a pixel and its nearest neighbor interaction – a 3x3 pixel array centered at the central pixel position $(x, y)$. (All information measures defined earlier are actually assigned to the central pixel of an arrangement. That was the reason why in a prior publication we call $I_{loc}$ the "Single Pixel Information Content", [25].)

The measure for intensity information $I_{int}(x, y)$ can be estimated as the mean absolute difference (MAD) between the central pixel gray level ($g_c$) and the gray levels of its 8 neighbors ($g_n$). Only $n$ results that are greater than zero are taken into account. Then the intensity component of information content can be expressed as:

$$I_{int}(x, y) = \frac{1}{n} \sum_{1}^{n} |g_c - g_n| \quad (3)$$

To account for topological information, the following procedure is applied: First, an expression for pixel's interrelationship with its surrounding is defined. We call it "status". It distinguishes between two perceptually discernable states: pixels that are at a lower intensity level than their surrounding neighbors and pixels that are equal or higher than their surrounding neighbors. Mathematically, status determination would be processed as:

$$stat = 8g_c - \sum_{i=1}^{8} g_i \quad (4)$$

Here $g_c$ is the gray level value of the central pixel and $\sum g_i$ is the sum of gray level values of its eight neighbors. The shortest status state description (encoding) would be in a binary form – "zero", if the subtraction result is negative, and "one" otherwise. Status is evaluated for every pixel in an image and mapped into an image-size status map.

Now, spatial (topological) interactions of a pixel with its nearest neighbors can be estimated using this map:

$$I_{top}(x, y) = p(1 - p) \quad (5)$$

Where $p$ is the probability that the central pixel and the surrounding ones share the same status (state). Because the support for each $I_{top}$ is defined as a 3x3 matrix, we can replace probability $p$ with the number of neighboring pixels $m$ that share the same status with the central pixel. The "1", correspondingly, must be replaced with the total number of the pixels engaged, which in our case (a 3x3 matrix) is equal to 8. The equation (5) can be then rewritten:

$$I_{top}(x, y) = m(8 - m) \quad (6)$$

$I_{top}$ values are computed for each pixel and saved in a special map of the size of the original image. At this time, using the two intermediate maps, the final $I_{loc}$ result can be computed and saved (mapped). Local image information content measure is thus evaluated and saved for further usage.

## 5.3 Experimental results

A possible effective use of local information content measure (descriptor) is elucidated by the following example. It is clear from the previous discussion, that peaks or local extremes in $I_{loc}$ may be seen as signs of a visible edge present at a given location. However, establishing a proper threshold for local extremums was always a hard and a sophisticated matter.

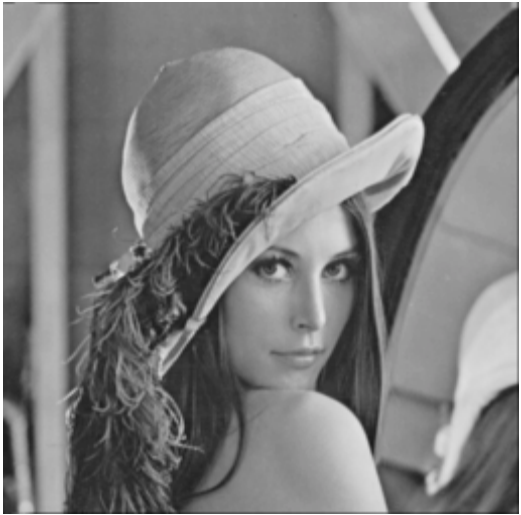

Fig. 8. The original Lena image.

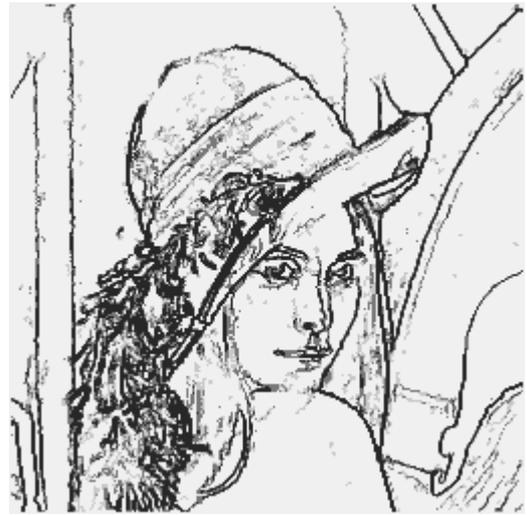

Fig. 10. Perceptually important (information rich) image locations.

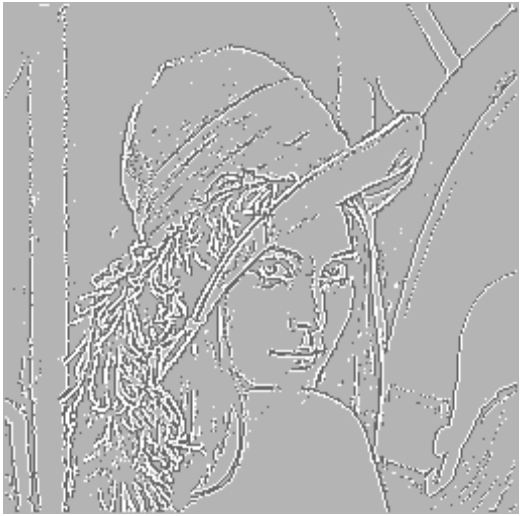

Fig. 9. Edge-localization image map.

Some examples of low-level (local) image information content evaluation.
Fig. 7 is the original image, 256 gray levels, 256x256 pixels size.
Fig. 8 shows the edge-localization map, where dark-gray is assigned to lower intensity sides of the edges and light-gray to the higher intensity edge sides.
Fig. 9 presents the information rich image points. In dark-gray are marked the most prominent image parts, carrying more than 50% of the total information content, half-gray is assigned to less important parts, carrying between 50 and 70% of information, in light-gray are marked the lowest rank importance parts, carrying between 70 and 85% of information content.

To overcome this difficulty, we propose to gather a cumulative histogram of $I_{loc}$ values. At first, the mean $I_{loc}$ value, estimated during $I_{loc}$ computations over the entire image, is multiplied by 3 and divided into a number of equal intervals (bins), which form the horizontal histogram axis. Then a histogram is constructed (by searching the whole $I_{loc}$ map) in accordance with the following rule: if the current $I_{loc}$ value is greater/equal than the bin's lower bound, its value is accumulated into this bin counter. As a result, the first bin represents the sum of all local information content measures, and histogram normalization can be easily accomplished. It is now explicitly visible what part of the whole "image information content" is carried out by $I_{loc}$ values that are equal/greater than a particular bin lower bound. This can be used as a threshold for appropriate image points assignment (marking). In such a way, a set of different information content related thresholds can be established, which can address diversified task related requirements. Fig. 10 shows Lena image, marked in this manner: the most prominent image points are marked in dark-gray, carrying more than 50% of the whole (low-level) information content. Less important image parts are marked in half-gray, carrying between 50% and 70% of information content and the lowest importance image parts are marked in light-gray, carrying 70% to 85% residuals of the information content.

The proposed technique can be effectively used to create more enhanced low-level information content descriptors. Since an edge is an abstraction, a symbolic line placed between two image parts that belongs to neither of them, the common practice of setting up one-pixel-wide edge marking lines is always a headache and a dilemma. Contemporary edge localization is problematic and controversial. On the contrary, $I_{loc}$-based edge localization certainly outlines edges as two coexisting, closely spaced line pair. Since edge is an intensity gradient, a sign of lower/higher gradient side (already available from the status map of $I_{top}$ computations) can be attached to each of the lines. This way, the edge localization problem receives a more easily achieved solution. A new sort of edge description appears, with a higher descriptive capacity. An example of double-line edge mapping can be found in Fig. 9.

## 6. Conclusions

We presented a new technique for unsupervised image information content generation and top-down image decomposition to its constituent visual sub-parts. Unlike traditional approaches, which adhere to bottom-up strategies, we propose a hybrid bottom-up/top-down strategy which produces the simplest (the shortest, in terms of Kolmogorov's Complexity) description of image information content. The level of unveiled description details is determined by the structures discernable in the image data and, thus, is independent from user requirements. We are aware that some of our assumptions and, consequently, the proposed solutions disagree with currently accepted theories and approaches in computer vision. So, we have provided evidence for their biologically plausibility.

Despite a seeming similarity to the established multimedia content description standards, which (like MPEG-7 standard, e.g.) provide means and rules for image information content creation and Schemas for Object Description Design, our proposed approach is principally different:

- MPEG-7 description creation relies on a bottom-up process, [33]. This poses extreme difficulties for the initial object segmentation/identification. Therefore such a task is left beyond the standard's scope.
- MPEG-7 is not supposed to provide image reconstruction from the descriptions. Analogously designed descriptors can only be used for image comparison and similarity investigation purposes, (such as in Content Based Image Retrieval and other Web-related applications, [34]).

With respect to the standardized techniques, our approach has palpable advantages. We provide a technique that autonomously yields a reasonable image decomposition (to its constituent objects), accompanied by concise object descriptors that are sufficient for reverse object reconstruction at different detail levels.

## Acknowledgements

The author would like to express his gratitude to the anonymous reviewers for their comments and remarks, which helped to improve the readability of this paper.